\documentclass[letterpaper, 10 pt, journal, twoside]{IEEEtran}

\usepackage{graphics} 
\usepackage{epsfig} 
\usepackage{mathptmx} 
\usepackage{times} 
\usepackage{amsmath} 
\usepackage{amssymb}  
\usepackage{algorithm}
\usepackage{algpseudocode}
\usepackage{multirow}
\usepackage{booktabs}
\usepackage{caption}
\usepackage{subcaption}
\usepackage{enumitem}
\usepackage[usenames,dvipsnames]{color}
\usepackage[noadjust]{cite}
\usepackage{soul}
\usepackage{color}
\usepackage{xcolor}
\usepackage{esvect}

\newcommand{\hide}[1]{{}}

\usepackage{marginnote}

\begin{document} 
\title{DeRi-IGP: Learning to Manipulate Rigid Objects \\Using Deformable Linear Objects via Iterative Grasp-Pull}

\author{Zixing Wang and Ahmed H. Qureshi
\thanks{Manuscript received: Oct, 26, 2024; Revised Jan, 5, 2024; Accepted Jan, 29, 2025.} 
\thanks{This paper was recommended for publication by Editor Tetsuya Ogata upon evaluation of the Associate Editor and Reviewers' comments.}
\thanks{The authors are with the Department of Computer Science, Purdue University, IN, USA.
        {\tt\small \{wang5389, ahqureshi\}@purdue.edu}}%
}

\let\oldtwocolumn\twocolumn
\renewcommand\twocolumn[1][]{
    \oldtwocolumn[{#1}{
    \begin{center}
            \includegraphics[width=\linewidth]{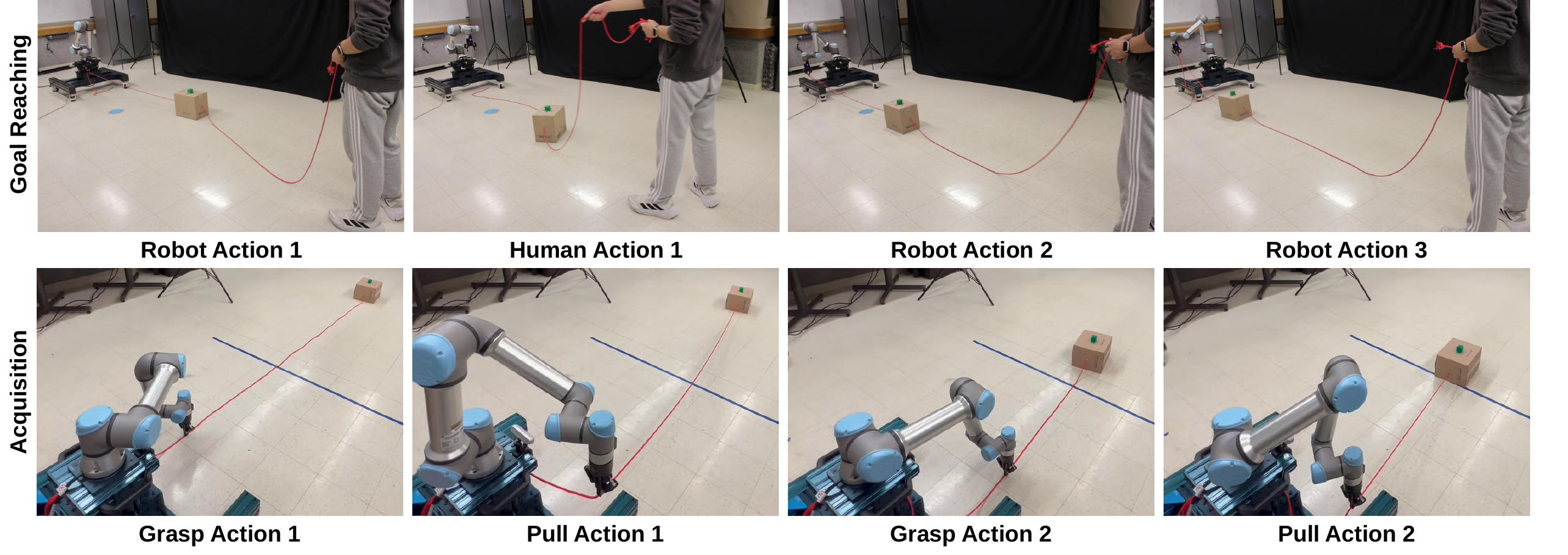}
            \captionof{figure}{DeRi-IGP utilizes an iterative grasp-pull moving primitive, which offers a large action and operational space, to carry out collaborative goal-reaching and long-distance object acquisition tasks. In the top row, we see DeRi-IGP performing the goal-reaching task, where it relocates a rigid object to a specified goal using ropes while working alongside a human collaborator. The bottom row illustrates the long-distance object acquisition task, in which the robot iteratively selects the grasping point on the rope based on local perception and identifies the corresponding pulling location to retrieve a distant rigid object.}
            \label{fig:title}
    \end{center}
    }]
}

\markboth{IEEE Robotics and Automation Letters. Preprint Version. Accepted Jan, 2025}
{Wang \MakeLowercase{\textit{et al.}}: DeRi-IGP}

\maketitle


\begin{abstract}

Robotic manipulation of rigid objects via deformable linear objects (DLO) such as ropes is an emerging field of research with applications in various rigid object transportation tasks. A few methods that exist in this field suffer from limited robot action and operational space, poor generalization ability, and expensive model-based development. To address these challenges, we propose a universally applicable moving primitive called Iterative Grasp-Pull (IGP). We also introduce a novel vision-based neural policy that learns to parameterize the IGP primitive to manipulate DLO and transport their attached rigid objects to the desired goal locations. Additionally, our decentralized algorithm design allows collaboration among multiple agents to manipulate rigid objects using DLO. We evaluated the effectiveness of our approach in both simulated and real-world environments for a variety of soft-rigid body manipulation tasks. In the real world, we also demonstrate the effectiveness of our decentralized approach through human-robot collaborative transportation of rigid objects to given goal locations. We also showcase the large operational space of IGP primitive by solving distant object acquisition tasks. Lastly, we compared our approach with several model-based and learning-based baseline methods. The results indicate that our method surpasses other approaches by a significant margin. The project supplementary material is available at: https://sites.google.com/view/deri-igp
\end{abstract}



\IEEEpeerreviewmaketitle

\section{Introduction}
\label{sec:intro}

\IEEEPARstart{T}{he} heterogeneous system manipulation task requires one or more robots to manipulate a rigid object via the connected deformable (soft) objects~\cite{wang2023deri}. Such configuration applies to various scenarios, such as transporting payloads with cargo sleds in snowy terrains and hauling tree chunks using ropes in forestry. The soft bodies in these systems provide enhanced maneuverability and portability because of their high degree of freedom. Consequently, we believe it is important to develop frameworks that can operate robots under such a heterogeneous configuration, leading to improvements in their capabilities and expanded application across a wide range of object transportation tasks.

Research in the field of heterogeneous system manipulation is still in its early stages and faces various challenges. Previous approaches use either model-based~\cite{donald2000distributed,corke2000experiments,maneewarn2005mechanics} or model-free~\cite{wang2023deri} methods to perform object repositioning using DLOs. Model-based methods~\cite{donald2000distributed,corke2000experiments,maneewarn2005mechanics} require hand-engineering, which limits their generalization across multiple agents, DLO types and rigid object shapes. Additionally, hand-engineering in these settings is complex because it requires determining the starting position where the rope begins to tighten for pulling, computing the rope's catenary curve, and estimating the rope's length precisely from raw perception. On the contrary, model-free methods learn from demonstration data to capture the underlying dynamics governing a given heterogeneous system. The development of model-free approaches remains limited, with only one approach available that is DeRi-Bot~\cite{wang2023deri}. DeRi-Bot learns from demonstration data and solves the rigid-object repositioning tasks using DLO. However, DeRi-bot operates with a top-down view of the entire scene and provides a pulling primitive under the assumption that the DLO is fully stretchable within robot reach and is already pre-attached to the robot hand. Thus, DeRi-Bot requires human intervention in case the rope slips out of the robot's hand and is also unable to handle large rope lengths and local field of view. 

To resolve the aforementioned problems of the existing model-free method, we propose DeRi-IGP, a new vision-based, model-free neural policy that uses \textbf{De}formable linear objects to manipulate \textbf{Ri}gid objects through parametrizing the Iterative Grasp and Pull (\textbf{IGP}) moving primitive. Our neural policy takes the environment perception from the robot's onboard sensor and outputs the robot's grasping and pulling position for the IGP primitive to manipulate the heterogeneous system. The process is repeated until a given task is achieved. Our IGP moving primitive is inspired by the process of individuals drawing water from a well — alternately using the left and right hand to grasp and pull the rope. Similarly, IGP encapsulates the iterative procedure of grasping a point on a rope, pulling it to a destination, and subsequently repeating the process until a given task is achieved. By imitating such an action model, DeRi-IGP presents the following advantages over prior DeRi-Bot. Firstly, it offers improved generalization across various rope lengths. Since the effective rope length is determined by the grasping point the robot selects, our framework seamlessly accommodates any rope of reasonable length without requiring modifications. In contrast, DeRi-Bot is constrained to fixed-length ropes, necessitating retraining whenever new rope lengths are introduced. Secondly, DeRi-IGP has a more extensive operational space. Since robots can execute multiple pulls, they can manipulate objects across a broader range of workspace, contingent upon its visibility within the camera's field of view, compared to scenarios where only a single movement is feasible. Lastly, our vision-based neural DeRi-IGP utilizes only the robot's onboard perception sensors without requiring a top-down view of the scene, allowing for generalizability across arbitrary robot positions. In summary, the main contributions of this work are summarized as follows:
\begin{itemize}
    \item We propose iterative grasp-pull, IGP, a universally applicable moving primitive for the heterogeneous system manipulation task.
    \item We design a DeRi-IGP framework that uses IGP and enables manipulation of rigid objects via deformable ropes with generalization ability across various rope lengths and rigid object sizes.
    \item We use a learning-based IGP action generation module to interpret the pattern governing the relationship between heterogeneous body systems and IGP actions.
    \item We employ a learning-based residual action-outcome prediction module to forecast position changes of rigid objects following the execution of IGP actions. Such a residual prediction scheme helps overcome problems caused by the inherent stochasticity of ropes, facilitates in the selection of the best IGP action from proposed options.
\end{itemize}
We demonstrate our method in both simulation and real-world scenarios by solving two tasks: multi-agent repositioning of rigid objects using DLOs, and signal agent acquisition of distant rigid objects with arbitrarily large DLOs. In the first task, we also show that our method can work alongside human partners to successfully reposition rigid objects using DLOs. We compare our method against various model-based and model-free methods. Our results indicate that our approach outperforms existing methods and exhibits one-shot generalization to real-world settings.

\section{Related Works}
\label{sec:related}
Soft-rigid body systems manipulation represents an emerging and relatively unexplored research area within the field of robotics. To the best of our knowledge, only a few previous works have explicitly addressed the problem of heterogeneous manipulation. As a result, we have expanded our related work discussion beyond heterogeneous systems and have explicitly included prior work related to rope manipulation due to their relevance to our proposed framework.

\subsection{Soft-rigid Body System Manipulation}
The seminal work of the soft-rigid body manipulation task is~\cite{donald2000distributed}. It proposes a model-based manually designed framework for packing and moving multiple objects via ropes. Then,~\cite{corke2000experiments,maneewarn2005mechanics} conducted more extensive research and analysis in this direction. In contrast, DeRi-Bot~\cite{wang2023deri} is a data-driven model-free method. It aims to move a rigid body block to the target position by controlling one or more robotic arms to pull the connected ropes. DeRi-Bot models the complex dynamics of soft-rigid body systems with neural networks. 
However, as mentioned in Section I, the DeRi-bot does not consider the robot grasping primitive, lacks generalization across different DLO lengths, and requires global information with a top-down field of view. 

\subsection{Soft Body Manipulation}
Model-based approaches in this field model DLOs' dynamics using various methods including implicit Finite Element Method (FEM)~\cite{du2021diffpd}, Material Point Method (MPM)~\cite{huang2019neural}, compliant Position-based Dynamics (XPBD)~\cite{10093017} and etc. Since we focus on model-free approaches for DLO manipulation such as ropes, this section is dedicated to that area of research. For an extensive survey on the robotic manipulation and sensing of deformable objects, please refer to~\cite{sanchez2018}.

In recent years, learning-based methods~\cite{nair2017combining,pathakICLR18zeroshot,schulman2016learning,sundaresan2020learning,wang2019learning,yan2020self,yan2021learning,chi2022irp,zhang2021robots} have grown rapidly and demonstrated impressive performance and generalization abilities. These methods have been demonstrated in various rope manipulation tasks such as reaching target DLO configuration, vaulting, knocking, and weaving. Regarding the target configuration reaching task,~\cite{nair2017combining} learns by imitating human demonstration, whereas~\cite{wu2020} uses a model-free reinforcement learning framework. Such tasks pay attention to soft objects' static configuration after manipulations. Conversely, dynamic manipulation tasks require an object to reach a certain configuration during the movement, which requires modeling the action-driven rope dynamics. In this dynamic manipulation domain, the vaulting, knocking, and weaving tasks were recently introduced and solved using a self-supervised framework in~\cite{zhang2021robots}. To make more precise dynamics control, the iterative residual policy (IRP)~\cite{chi2022irp} is proposed. It leverages the residual physics~\cite{zeng2020tossingbot} and the Delta Dynamics Network to iteratively refine the generated actions for various types of tasks, such as rope whipping that requires the rope end to reach certain positions. Our work also chooses the learning-based direction to overcome the complexity of the analytical physics modeling of DLOs and their interaction with rigid objects. 

\begin{figure*}[t]
  \centering
  \vspace{3mm}
  \includegraphics[width=\textwidth]{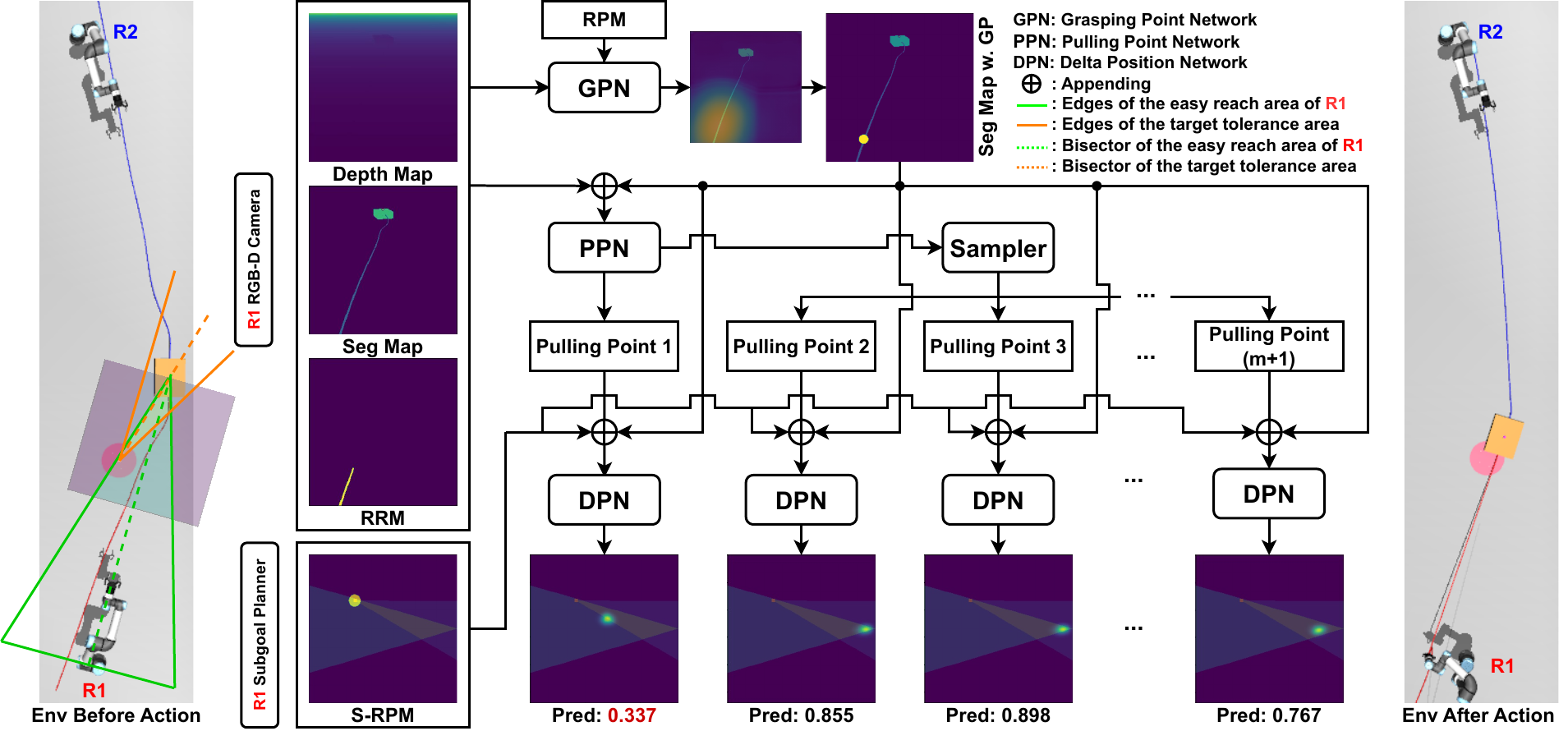}
  \caption{
  In the DeRi-IGP process, the subgoal planner assigns local targets to agents, while the GPN and PPN predict grasping and pulling points based on observations. These predicted actions are then used by the DPN to forecast the object's future position and select the best action that minimizes the object-target distance. It's important to note that all proposed robot actions will be compared for selection. The subgoal planner's map presented in the left-most sub-figure is also used as a mask on the spatial relative position map and the output of DPN.
  }
  \label{fig:workflow}
  \vspace{-0mm}
\end{figure*}

\section{Methods}
\label{sec:methods}
This section presents our DeRi-IGP framework, its core components, and their synergy to solve heterogeneous system manipulation tasks. 
\subsection{State-space Representation}
An instance of the experimental environment contains $n\in \mathbb{N}$ number of agents equipped with RGB-D sensors, a rigid object connected with the $n$ number of DLOs (ropes), and a position indicator placed at the center of the top surface of the rigid object. We use the following notation to represent their states. Let the state of $n$ agents be denoted as $C = \{c^1,~c^2,~\dots,~c^n\}$. Let $D = \{d^i,~d^2,~\dots,~d^n\}$ denotes the depth map of the task environment captured by each agent, where $D \subset \mathbb{R}^{w \times w}$ and $w$ represents the map side length. Let $S = \{s^1,~s^2,~\dots,~s^n\}$ denotes the segmentation maps corresponding to $D$, where $S \subset \{0,~1,~2,~3,~4\}^{w \times w}$, in which 1, 2, 3, 4 and 0 represent the rope, object, position indicator, grasping point and all other elements, respectively. To help identify the reachable part of a rope, the binary Rope Reachability Maps (RRMs) are generated using depth map for each agent and denoted as $R = \{r^i,~r^2,~\dots,~r^n\}^{w \times w}$, where $R \subset \{0, 1\}$, in which 1 and 0 represent graspable part of the rope and all other areas, respectively. Specifically, in a $r$, a pixel is marked as unreachable if the distance is longer than our robot arm’s max extending length. In addition, we use Spatial Relative Position Maps (S-RPM) denoted as $P=\{p^1,~p^2,~\dots,~p^n\},~P\subset\mathbb{R}^{w \times w}$ to depict the target position with respect to the rigid object. Similar to~\cite{wang2021spatial,chen2023efficient,wang2023deri}, it indicates the target position in a 2D array that the rigid object's position always resides in the middle of the right edge, as also indicated in Fig.~\ref{fig:workflow}. In addition, we encode the same information with an alternative format, Relative Position Map (RPM) $V=\{v^1,~v^2,~\dots,~v^n\},~V\subset\mathbb{R}^{w \times w}$. This map is produced by broadcasting and padding the two-dimensional vector representing the position disparity. Further insights into their application and analysis are elaborated upon in Section~\ref{subsec:gpnppn}.

\subsection{Action-space Representation}
An IGP action denoted as $\phi^i \in \Phi$, performed by an agent $c^i$, is defined by a grasping point $\lambda^i \in \Lambda$ and a pulling point $\eta^i \in H$: $\phi^i = (\lambda^i,~\eta^i)$,  where $\Lambda, H\subset \mathbb{R}^3$. Given an IGP action, the agent approaches the grasping point and grasps the rope, then moves to the pulling point following the shortest linear path.

\subsection{Action Generation Module}
\label{subsec:gpnppn}
An IGP action comprises a grasping and a pulling point. Our action generation module produces IGP actions through the following networks and a sampling approach.

\textbf{Grasping Point Network (GPN)}. It employs a neural network to determine the suitable grasping point based on the prevailing environmental state. Specifically, for agent $c^i$ at time step $t$, the input for GPN is a 4-channel 2D array. It comprises the depth map $d_t^i$, the segmentation map $s_t^i$, the RRM $r_t^i$ and the RPM $v_t^i$:\vspace{-0.05in}
\[
\lambda_t^i = \text{GPN}(d_t^i,~s_t^i,~r_t^i,~v_t^i),
\vspace{-0.05in}\]
where $s_t^i \subseteq \{0,~1,~2,~3\}$. The last channel $v_t^i$ is formed by broadcasting and zero-padding a 4D vector, in which the first and the last two numbers are the X-Y coordinate of the object and the target within $c^i$'s frame. The output $\lambda_t^i$ is a 2D probability map of the same dimension as $d_t^i$ defined in the same domain. From this map, we sample the reachable grasping point. Finally, given the predicted grasping point, $c^i$ will move the end-effector to the corresponding position and grasp the rope. Please note that we assume ropes always reside on the ground plane, so the heights of all the grasping points are identical and fixed.

\textbf{Pulling Point Network (PPN)}. It uses a neural network to predict the proper pulling point. To be more specific, PPN takes as input the depth map $d_t^i$, the segmentation map $s_t^i$ with $\lambda_t^i$ encoded, and the S-RPM $p_t^i$:
\[\vspace{-0.05in}
\eta_t^i = \text{PPN}(d_t^i,~s_t^i,~\lambda_t^i,~p_t^i),
\]
where $s_t^i \subseteq \{0,~1,~2,~3,~4\}$. It encodes the same information as $s_t^i$ used in GPN along with the grasping point positions. The output $\eta_t^i$ is a 3D vector, indicating the destination position of the IGP action. 

Note that GPN and PPN use different formats to represent the relative position between the target and the rigid object. 
According to our experiment result, the performance of GPN with S-RPM, $p^i_t$, is worse than our presented design. We conjecture the reason is that since $\lambda_t^i$, $d_t^i$ and $s_t^i$ are defined in the RGB-D camera's view (pixel-aligned), while $p^i_t$ is defined in a synthesized top-down view. This divergence might mislead the neural network to interpret the goal position in $p^i_t$ as if it were in the camera's viewpoint.

\textbf{Action Sampling Strategy}. Soft body dynamics exhibit high complexity and sensitivity to perturbations. A minor change in action can lead to significantly divergent outcomes. Our model predictive scheme addresses this uncertainty by evaluating multiple actions distributed around the neural network's initial prediction. It provides a probabilistic approach to action selection, improving robustness beyond a single network prediction. Specifically, given the predicted pulling point~$\eta_t^i$ of the IGP action $\phi_t^i = (\lambda_t^i,~\eta_t^i)$, we sample $m$ extra IGP actions $\hat{\Phi}_t^i = \{(\lambda_t^i,~\eta_t^{i:1}),~(\lambda_t^i,~\eta_t^{i:2}),~\dots,~(\lambda_t^i,~\eta_t^{i:m})\}$ from a Gaussian distribution centered at $~\eta_t^i$ with a predefined variance 0.5 $meter^2$, i.e., $\hat{\Phi}_t^i \sim \mathcal{N}(\phi_t^i,~0.5)$.

\subsection{Residual Action-Outcome Prediction Module}
\label{subsec:dpn}
The residual action-outcome prediction module anticipates the outcome rigid object position after executing an IGP action. Such a function is necessitated by the action sampling strategy of the framework. We build the Delta Position Network (DPN) to perform the task.

\textbf{Delta Position Network}. Given a series of predicted and sampled IGP actions $\{\phi^i_t,~\hat{\phi}^{i:1}_t,~\hat{\phi}^{i:2}_t,~\dots,~\hat{\phi}^{i:m}_t\}$ for $c^i$ at step $t$, we use DPN to forecast future state of the rigid object. Each time DPN takes as input the depth map $d_t^i$, the segmentation map $s_t^i$, and an IGP action, $\phi_t^i$ or $\hat{\phi}^{i:m}_t$, to predict the rigid object's new position with respect to its original position:
\[
\Tilde{p}_t^i = \text{DPN}(d_t^i,~s_t^i,~\phi_t^i)~or~\text{DPN}(d_t^i,~s_t^i,~\hat{\phi}^{i:m}_t)
\]
where $\Tilde{p}_t^i$ is a S-RPM of the same format as $p$ used in PPN. Among the IGP actions of all the agents, the one predicted to result in the greatest reduction of distance between the rigid object and the target will be selected to execute. The neural network structure of DPN is exactly the same as GPN.

\begin{figure*}[t]
  \centering
  \vspace{3mm}
  \includegraphics[width=1.\textwidth]{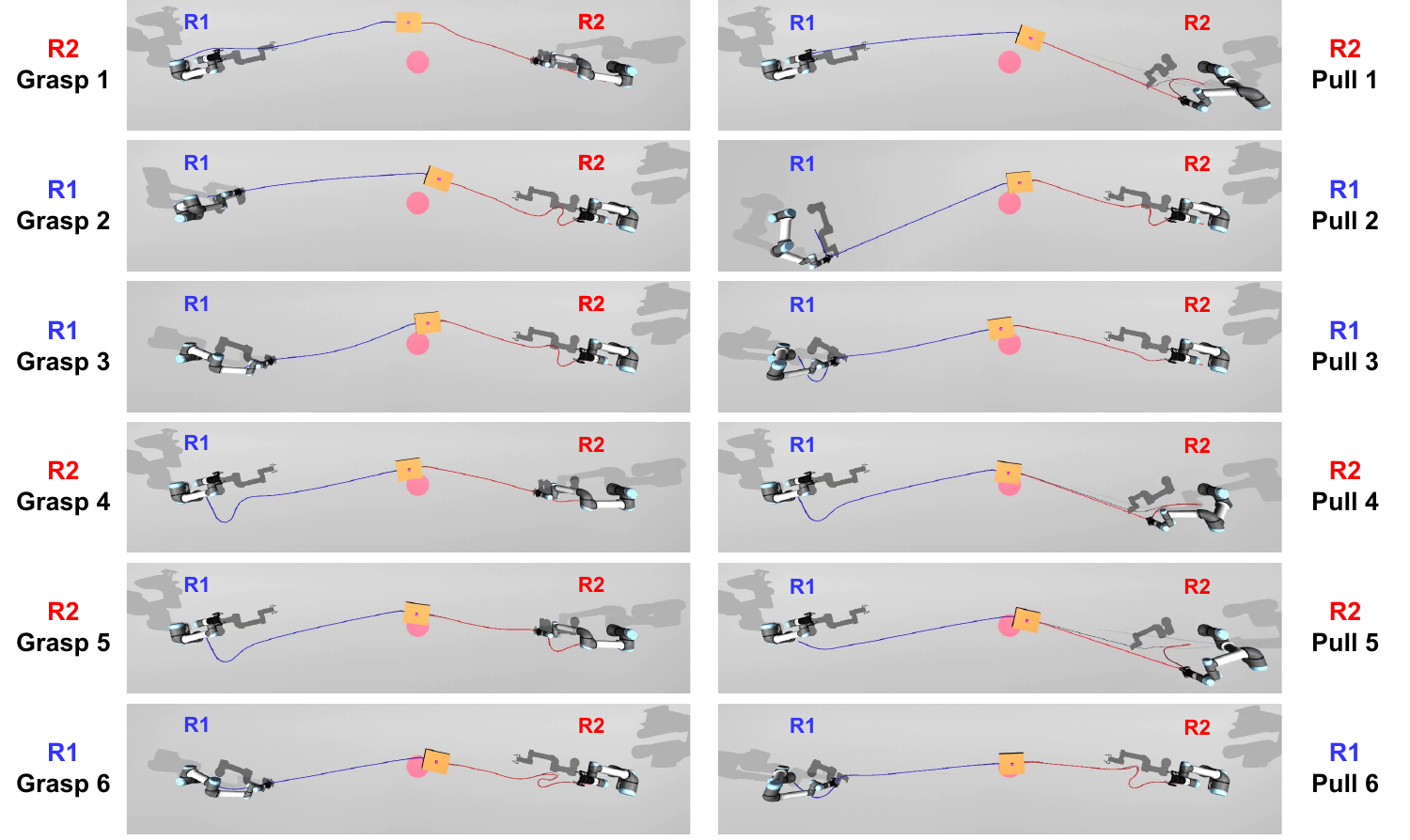}
  \caption{An example of dual-bot random-position goal-reaching task. After six IGP actions, the rigid object (brown block) is moved to the target position (pink circle). Note that the task configuration is under-actuated given the relative position between the rigid object and the target, and yet DeRi-IGP finishes the task with multi-robot cooperation}
  \label{fig:sim_exp}
  \vspace{-0mm}
\end{figure*}

\subsection{Subgoal Planner}
\label{subsec:subgoal}
Because of its iterative nature, DeRi-IGP has an extensive operational space, so directly employing the target position as the objective at every step presents two challenges. First, since the S-RPM size is fixed and limited, it is impossible to represent a very far position within the map. Second, since all our neural networks use 1-step action-outcome pairs for training, distant targets may be out-of-distribution. To resolve these problems, we introduce two types of subgoal planners.

(1) \textbf{Linear Greedy Planner (LGP)}. LGP determines the subgoal position as the target position if it resides within the planning window. Otherwise, LGP uses the point of intersection between any edge of the current S-RPM and the line segment linking the target and rigid object's position instead. (2) \textbf{Geometric Intersection Planner (GIP)}. GIP determines the valid subgoal region given two areas. First, an isosceles triangular defining easy-reach area with a base perpendicular to the line segment between the agent and the rigid object position. The length of the base is twice the maximum reaching length. We use a triangle to represent this area due to our linear moving primitive. Second, the target tolerance area, which is a circular sector centered at the object's position. The sector has an angular span of $\pi/6$ and an infinite radius. Its bisector line passes through the object's current and target location. Within the part of the intersection of the two areas falling inside the corresponding map $p$, the planner picks the point nearest to the target position as the subgoal, illustrated in Fig.~\ref{fig:workflow}. 

\subsection{Multiagent Workflow}
\label{subsec:workflow}

The workflow of DeRi-IGP is identical to DeRi-Bot~\cite{wang2023deri}. In an episode, at each step:
\begin{itemize}
    \item GPN and PPN collaboratively propose an IGP action for each of the $n$ agents.
    \item The Gaussian sampler generates $m$ additional commands around each proposed action, leading to a total of $n \times (m+1)$ samples.
    \item The generated IGP actions are then processed by DPN to forecast rigid-body positions in the environment.
    \item Finally, based on DPN output, we select the best action and its associated robot for execution, yielding the shortest distance $l$.
\end{itemize}
This process is repeated until termination. To avoid object sliding and drifting, we limit the max velocity of the end-effector.

\subsection{Neural Network Architecture}
We use different neural network architectures for GPN, PPN, and DPN. Regarding PPN, we leverage the ResNet-34 variant of ResNet~\cite{he2016deep} followed by a two-layer MLP~\cite{rosenblatt1958perceptron} network to predict a 3-dimensional vector. PPN and DPN use Deeplab V3 plus~\cite{chen2018encoder} structure, which has been widely used in generating probability masks.

\subsection{Data Generation}
\label{subsec:datagen}
We use MuJoCo~\cite{todorov2012mujoco} to build our simulated setup. The data generation process of DeRi-IGP is similar to~\cite{wang2023deri}. However, we employ a single-robot setup to accommodate the extensive operational space. Additionally, we limit the max velocity of the robot end-effector to avoid sliding and drifting. It is important to emphasize that in this study, we deliberately constrain the action speed of agents to uphold a quasi-static process. Consequently, acceleration in the process is negligible since the moving speed of the object is slow. Our setup necessitates frequent resets of the data collection environment, placing the rigid object at random distant positions and randomizing the environment configurations. We randomize the dimension of the rigid object for generalization. For each move, we sample and execute $\phi$ and collect the initial and final depth map, the color map, and the rigid object's positions. After processing, we form an instance of the dataset, i.e., ($\phi_t$, $d_t$, $s_t$, $r_t$, $v_i$, $p_i$). We collected 20,000 instances and split them into training, validation, and testing with 0.8: 0.1: 0.1 ratios.

\subsection{Training and Testing Details}
\label{subsec:td}
We construct all our networks using PyTorch-Lightning~\cite{falcon2019pytorch} and train them using AdamW optimizer~\cite{Loshchilov2017DecoupledWD}. 
The final MSE loss of PPN on the training and the test set is $0.021$ and $0.024$ meters. As for GPN and DPN, their final BCE losses on both the training and test sets were $1.06\times10^{-4}$, $1.17\times10^{-4}$, and $1.06\times10^{-4}$, $1.18\times10^{-4}$.

\section{Experiments}
\label{sec:experiments}
This section presents our evaluation analysis for DeRi-IGP in solving long-distance and goal-reaching acquisition tasks. In all experiments, we vary the rope length and object dimensions randomly. The box dimensions are sampled uniformly within the following ranges, measured in meters: the depth ranges from 0.10 to 0.15, the height ranges from 0.05 to 0.10, and the width ranges from 0.10 to 0.20. The length of the rope is uniformly sampled between 2.5 and 4.0 meters. In both the simulator and real world, we employ realsense D455 cameras and UR5e arms with its built-in positional controller to conduct the experiments.
\subsection{Comparison Methods}
\label{subsec:cand}
We consider the following methods for comparison:
\begin{itemize}
    \item \textbf{DeRi-IGP-GIP}: This method uses the DeRi-IGP framework with GIP as a subgoal planner.
    \item \textbf{DeRi-IGP-LGP}: This method uses the DeRi-IGP framework with LGP as a subgoal planner. 
    \item \textbf{DeRi-IGP-NS}: This baseline operates similarly to DeRi-IGP-GIP but without the action sampling process facilitated by DPN. In this context, DPN is solely employed for evaluating IGP actions generated by neural networks.
    \item \textbf{DeRi-IGP-RG}: This baseline operates similarly to DeRi-IGP-GIP but without GPN. The grasping point is randomly sampled from the corresponding RRM.
    \item \textbf{DeRi-Bot}~\cite{wang2023deri}: The predecessor of DeRi-IGP. It employs a sampled-based action proposal framework similar to DeRi-IGP. However, it assumes a pre-grasped rope.
    \item \textbf{Tension-driven Approach (TDA)}: This model-based baseline grabs and lifts the rope to the height of the robot's base to maximize movement range. It then selects a pulling path that is parallel and equal in length to the object-to-target vector.
    \item \textbf{Goal-directed Tension-driven Approach (G-TDA)}: G-TDA tries to align the pulling path with the object-to-target vector. After lifting the rope, G-TDA moves it to the nearest point on the object-to-target vector and then pulls the rope along that direction.
\end{itemize}
Note, we set $m = 10$ for all the methods with the action sampling process.

\subsection{Metrics}
\label{subsec:metrics}
We use the following metrics for evaluation:
\begin{itemize}
    \item \textbf{Success Rate (SR)}: This presents the percentage of successful versus total episodes based on the predefined success conditions of different tasks.
    \item \textbf{Shortest Offset (SO)}: This provides the historical shortest distance in meters between the object's current and target position during an episode.
    \item \textbf{Final Offset (FO)}: The final distance in meters between the object's current and target position.
    \item \textbf{Step Cost (SC)}: This indicates the number of actions (steps) executed during each episode. 
\end{itemize}
We report these metrics' mean and standard deviation, except for the success rate, across all successful test scenarios.

\begin{table}[t]
\vspace{2mm}
\caption{Evaluation results on a goal-reaching task}
    \begin{center}
    \vspace{-4mm}
    \resizebox{0.48\textwidth}{!}{
        \begin{tabular}{c | c c c c }
        \toprule
                     & SR $\uparrow$     & SO $\downarrow$     & FO $\downarrow$       & SC $\downarrow$ \\ \midrule
        DeRi-IGP-GIP & 86\%  & 0.139 $\pm$ 0.087 &\textbf{0.147} $\pm$ \textbf{0.088} & 4.320 $\pm$ 2.485 \\
        DeRi-IGP-LGP & \textbf{88\%}  & \textbf{0.130} $\pm$ \textbf{0.071}  & 0.215 $\pm$ 0.197  & 6.12 $\pm$ 2.233 \\
        DeRi-IGP-NS  & 48\%  & 0.247 $\pm$ 0.160  & 0.291 $\pm$ 0.178  & \textbf{3.540} $\pm$ \textbf{2.022}\\ 
        DeRi-IGP-RG  & 78\%  & 0.151 $\pm$ 0.097  & 0.177 $\pm$ 0.123  & 4.16 $\pm$ 1.943\\
        TDA          & 72\%  & 0.179 $\pm$ 0.145 & 0.193 $\pm$ 0.152  & 4.38 $\pm$ 1.896 \\ 
        G-TDA        & 72\%  & 0.203 $\pm$ 0.194 & 0.230 $\pm$ 0.248 & 5.26 $\pm$ 2.827 \\ \bottomrule
        \end{tabular}}
     \end{center}
     \label{tb1}
    \vspace{-3mm}
\end{table}

\subsection{Goal-reaching Task}
\label{subsec:gr}
In a goal-reaching task episode, two agents are required to work collaboratively to move the rigid object to the target position using DLO. After the execution of an IGP action, the episode is terminated if: (i). if the rigid object is less than 0.1 meters away from the target, (ii) if the distance $l$ has consecutively increased over the past three steps compared to the lowest historical value during the episode, and (iii) if the ropes of all agents disappear from the sensor's view or if it only happens to one agent and the rigid object's position is closer to the other agent than the target position since we cannot push objects with ropes. Finally, a successful episode is one where the final distance $l$ is less than 0.2 meters.

\begin{table}[t]
\vspace{2mm}
\caption{Evaluation results on a fixed-position setup}
    \begin{center}
    \vspace{-4mm}
    \resizebox{0.48\textwidth}{!}{
        \begin{tabular}{c | c c c c }
        \toprule
                     & SR $\uparrow$  & SO $\downarrow$     & FO $\downarrow$     & SC $\downarrow$ \\ \midrule
        DeRi-IGP-GIP & \textbf{70\%}  & \textbf{0.186} $\pm$ \textbf{0.126}  &\textbf{0.202} $\pm$ \textbf{0.131}  & 5.620 $\pm$ 2.630 \\
        DeRi-Bot~\cite{wang2023deri} & 32\%  & 0.260 $\pm$ 0.179  & 0.269 $\pm$ 0.184  & \textbf{5.400} $\pm$ \textbf{1.822} \\ \bottomrule
        \end{tabular}}
     \end{center}
     \label{tb2}
    \vspace{-6mm}
\end{table}

\begin{figure*}[t]
  \centering
  \vspace{2.5mm}
  \includegraphics[width=\textwidth]{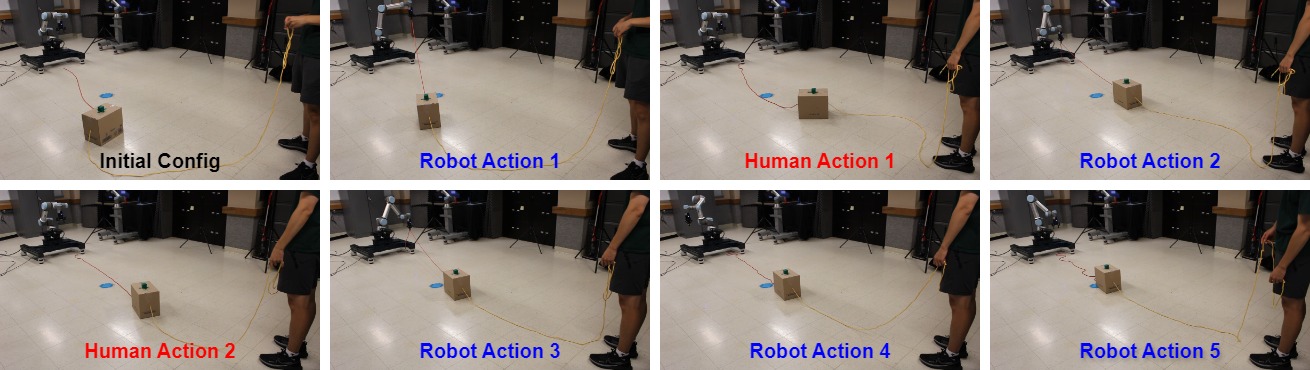}
  \caption{A demonstration of the long-horizon human-robot collaborative goal-reaching task. The human operator and the robot take turns manipulating the rope to move the object (brown card box) to the target position marked in blue. Note, the robot failed to pick up the rope in Action 4.
  }
  \label{fig:real_exp_grid}
  \vspace{-6mm}
\end{figure*}

\subsubsection{Simulated goal-reaching task evaluation}
\label{subsubsec:srpe} In each episode, we uniformly sampled the positions of the agents and the rigid object's initial and target locations within the valid range. The purpose of this evaluation is to (i) assess the manipulation capability of DeRi-IGP, (ii) check the necessity of a model-free approach, and (iii) validate the generalization ability across distinct environment configurations. 

Table~\ref{tb1} summarizes the results, accompanied by a visual depiction of a task in Fig.~\ref{fig:sim_exp}. It can be seen that the other methods outperform DeRi-IGP-NS by a considerable margin, providing a compelling rationale for the use of the DPN-based action sampling framework. The step cost of DeRi-IGP-NS is low because it solved fewer cases than other methods, mostly with a short horizon. Regarding grasping point selection, the relatively low performance of DeRi-IGP-RG indicates the importance of GPN. It also reflects the efficiency of PPN. Specifically, even with a random grasping point, PPN can still generate a reasonable pulling point to finish the task. In addition, DeRi-IGP-RG outperforms DeRi-IGP-NS, likely due to differences in action space complexity. While pulling points cover a broader range, nearly equivalent to the robot arm's operational space, grasping points are more constrained along the rope, which explains the varying performance and sensitivity of pulling and grasping point selections. Furthermore, by evaluating subgoal planners, the results suggest comparable performance levels among them. However, GIP notably takes fewer steps than LGP and demonstrates higher stability, as indicated by FO. In addition, we exclude PPN's ablation since in DeRi-Bot~\cite{wang2023deri}, the informed sampling and random sampling baseline for pulling action have already been shown to exhibit poorer performance than the data-driven neural network-based pulling approach. Finally, the two model-based baselines, TDA and G-TDA, exhibit relatively poor performance than our method. It is important to highlight that these model-based baselines directly utilize simulator feedback to determine both the pulling distance and direction. In contrast, all other DeRi-IGP baselines rely on relatively low-resolution depth images for spatial information. This outcome indicates that tension-driven, model-based methods, despite having accurate system information, struggle to solve the given task. Note that we exclude DeRi-Bot~\cite{wang2023deri} as it lacks grasping primitive and cannot generalize to varying rope lengths and environment size and setup.

\subsubsection{Simulated Fixed-position Evaluation}
\label{subsubsec:sfpe}
In this experiment, we compare our approach against Deri-Bot~\cite{wang2023deri}. We employ the dual-bot simulated evaluation procedure introduced in~\cite{wang2023deri}, i.e., we keep the agent position fixed and only randomize the rigid object initial and target location. Since the operational space of this task is smaller than the range of $p$, the GIP and LGP will generate the same goal positions. Thus, we only keep DeRi-IGP-GIP in this experiment. The objective is to perform a quantitative comparison between the two distinct DeRi-IGP and Deri-Bot moving primitives under identical settings. We report the related results in Table~\ref{tb2}. As the results indicate, DeRi-IGP demonstrates overwhelming advantages over DeRi-Bot. Since both methods operated under the same environment and leveraged the sampled-based action proposing module, we conclude that the significant disparity in performance is attributed to the IGP moving primitive.

\subsubsection{Real-world Human-Robot Evaluation}
We conduct the real-world goal-reaching task under the human-robot operational configuration\footnote{The human-robot collaboration study in this paper has been approved by the Purdue Institutional Review Board (Purdue IRB-2024-1299). We follow the related protocols to ensure the human subjects' privacy, confidentiality, and anonymity.}. An example of such a task is presented in Fig.~\ref{fig:real_exp_grid}. Such a task assesses the sim-to-real transfer capability and the human-robot team compatibility of DeRi-IGP. Note that DeRi-IGP exhibits an asynchronous behavior, so it is expected to naturally collaborate with human operators. In our experiment, the human operator holds the highest priority during task execution. The robot will only take action if the human operator does not take action within a specified time threshold.

We invited five volunteers to participate in our experiments. The participants have no knowledge about the system except the operational process to avoid bias. We asked them to use their own judgment to act to shrink the object-target distance. The quantitative results of the experiments are as follows. The average FO of all the experiments is 0.126, and the average SC 4.11. Such a result resembles the simulated results, which demonstrate that our framework can generalize to real-world environments and can collaborate with human partners without retraining or extra tuning. In terms of SC, we observed that sometimes the robot gripper failed to grasp the rope, which increased the cost. However, this issue does not affect SR since our framework can naturally resolve grasping failures by its iterative behavior. Apart from quantitative results, we conducted user study survey to collect subjective qualitative feedback. The questions for each participant and the corresponding results are presented as follows:
\begin{enumerate}
    \item Did robot actions aid in moving the object to the target? \\(Yes: 100\%, No: 0\%)
    \item Are the time and step costs acceptable for the task? \\(Yes: 100\%, No: 0\%)
    \item Did you find the robot conduct any irrational action that hindered the progress of moving the object to the target location? (Yes: 0\%, No: 100\%)
    \item Did you feel the robot required too much human intervention or guidance? (Yes: 0\%, No: 100\%)
    \item Was there any moment where you wished the robot should have acted differently? (Yes: 40\%, No: 60\%)
\end{enumerate}
According to the results, all participants affirmed that the robot's actions successfully aided in moving the object to the target, and every respondent also found the time and step costs acceptable. Notably, there were no reports of irrational actions hindering progress, underscoring the robot's reliability and functionality. It worth noting that though 40\% expressed that they wished the robot had acted differently at some moment, there were no reports of irrational actions hindering progress. We believe such a result suggests that the framework's performance was largely effective, while there is still room for improvement in gaining human trust.


\subsection{Distant Object Acquisition Task}
\label{subsec:doa}
This task requires the agent to relocate a distant, rigid object to a neighboring position using ropes. Such a task effectively verifies the claimed extensive operational space of the IGP moving primitive. For this task, the target position is consistently set 0.1 meters ahead of the agent. We use the termination criteria from Section V-C, except using a 0.5-meter threshold for the first condition. An episode is deemed successful when the final $l$ is under 0.5 meters.

\subsubsection{Simulated Evaluation}
\label{subsubsec:slda}
We conducted 10 episodes of this task within the simulator. In the $i^{th}$ episode, the rigid object is initialized at a random position, which can be seen by the RGB-D sensor, with a distance of $4.5 + i \times 0.5$ meters from the agent. Finally, DeRi-IGP achieved a 100\% success rate. Besides the broad operational space, we believe the results indicate our IGP action generation module consistently generates reasonable actions in long-horizon tasks.

\subsubsection{Real-world Evaluation}
\label{subsubsec:rlda}
In addition to the simulated experiment, we perform the distant object acquisition task in the real world to validate the sim-to-real generalization ability of our framework. An illustrative instance can be seen in Fig.~\ref{fig:title}. Across 10 episodes, we attained a 90\% success rate, with a single failure instance attributed to the rope getting entangled with the robot base.

\section{Conclusions and Future work}
\label{sec:con}
This work presents a universally applicable and effective moving primitive called Iterative Grasp and Pull for manipulating DLO. It also introduces a neural policy called DeRi-IGP, which parameterized IGP primitives for multiagents to solve the rigid object transportation tasks using DLO. Our results indicate that DeRI-IGP outperforms various model-free and model-based baseline methods and exhibits larger operational space and better generalization to various task configurations. 

In the future, we aim to scale our proposed framework to more complex tasks, such as rigid object rotation via soft bodies. We also aim to solve object transportation tasks in constrained environments with obstacles. Another possible avenue to extend our approach is to employ mobile manipulators, which will further enhance the operational space and allow more precise manipulation of heterogeneous systems. Additionally, we aim to explore how human predictive models can be incorporated into our framework to allow for synchronous human-robot collaboration for solving object transportation tasks using ropes. 


\bibliographystyle{IEEEtran}
\bibliography{root}

\end{document}